\documentclass{article}

\usepackage{microtype}
\usepackage{graphicx}
\usepackage{booktabs} 

\usepackage{hyperref}
\usepackage{url}
\usepackage{mathtools}
\usepackage{amsmath}
\usepackage{amsthm}
\usepackage{graphicx}
\usepackage{amssymb}
\usepackage{caption}
\usepackage{subcaption}
\usepackage{array}
\usepackage{enumitem}

\usepackage{xcolor}

\usepackage{hyperref}

\usepackage[accepted]{icml2018}

\icmltitlerunning{Selecting Representative Examples for Program Synthesis}

\DeclareMathOperator*{\argmin}{\arg\!\min}

\newcommand{\subD}{D'}
\newcommand{\D}{D}
\newcommand{\greedD}{D^{g}}

\newcommand{\prune}{prune}
\newcommand{\neib}{nb}
\newcommand{\rem}{rem}

\begin{document}

\twocolumn[
\icmltitle{Selecting Representative Examples for Program Synthesis}



\icmlsetsymbol{equal}{*}

\begin{icmlauthorlist}
\icmlauthor{Yewen Pu}{MIT}
\icmlauthor{Zachery Miranda}{MIT}
\icmlauthor{Armando Solar-Lezama}{MIT}
\icmlauthor{Leslie Pack Kaelbling}{MIT}
\end{icmlauthorlist}

\icmlaffiliation{MIT}{Massachusetts Institute of Technology}

\icmlcorrespondingauthor{Yewen Pu}{yewenpu@mit.edu}

\icmlkeywords{Machine Learning, ICML}

\vskip 0.3in
]



\printAffiliationsAndNotice{}  

\begin{abstract}
Program synthesis is a class of regression problems where one seeks a solution, in the form of a source-code program, mapping the inputs to their corresponding outputs exactly. Due to its precise and combinatorial nature, program synthesis is commonly formulated as a constraint satisfaction problem, where input-output examples are encoded as constraints and solved with a constraint solver. A key challenge of this formulation is scalability: while constraint solvers work well with a few well-chosen examples, a large set of examples can incur significant overhead in both time and memory. We describe a method to discover a subset of examples that is both small and representative: the subset is constructed iteratively, using a neural network to predict the probability of unchosen examples conditioned on the chosen examples in the subset, and greedily adding the least probable example. We empirically evaluate the representativeness of the subsets constructed by our method, and demonstrate such subsets can significantly improve synthesis time and stability.
\end{abstract}

\section{Introduction}

Program synthesis (or synthesis for short) is a special class of regression problems where rather than minimizing the error on an example dataset, one seeks an exact fit of the examples in the form of a program. Applications include synthesizing database relations \citep{RohitEMSynth}, inferring excel-formulas \citep{Flashfill}, and compilation \citep{PhothilimthanaT16}. The synthesized programs are complex, consisting of branches and loops. Recent efforts \citep{EllisST15,RohitEMSynth} show an interest in applying the synthesis technique to large sets of examples, but scalability remains a challenge. We present a method that selects a small \emph{representative subset} of examples from a dataset, such that it is sufficient to specify a correct program, yet small enough to encode efficiently.

There are two key ingredients to a synthesis problem: a domain specific language (DSL for short) and a specification. The DSL defines a space of candidate programs which serve as the model class. The specification is commonly expressed as a set of input-output examples which the candidate program needs to fit exactly. The DSL restricts the structure of the programs in such a way that it is impossible to fit the input-output examples in an ad-hoc fashion: This structure aids generalization to an unseen input despite fitting the training examples exactly. 

Given the precise and combinatorial nature of synthesis, gradient-descent based approaches perform poorly and an explicit search over the solution space is required \citep{Terpret}. For this reason, synthesis is commonly casted as a constraint satisfaction problem (CSP) \citep{sketch13,JhaGST10}. In such a setting, the DSL and its execution can be thought of as a parametrized function $F$, which is encoded as a logical formula. Its parameters $s \in S$ correspond to different instantiations of programs within the DSL, and the input-output examples $D$ are expressed as constraints which the instantiated program needs to satisfy, namely, producing the correct output on a given input. 
\begin{equation*}
\exists s \in S. ~~ \bigwedge_{(x_i,y_i) \in D} F(x_i;s)=y_i ~.
\end{equation*} 
The encoded formula is then given to a constraint solver such as Z3 \citep{z3}, which solves the constraint problem, producing a set of valid parameter values for $s$. These values are then used to instantiate the DSL into a concrete, executable program.

\paragraph{A key challenge} of framing a synthesis problem as a CSP is that of scalability. While solvers have powerful heuristics to efficiently prune and search the constrained search space, constructing and maintaining the symbolic formula over a large number of constraints constitutes a serious overhead\footnote{However, if the solver does manage to construct and maintain all the constraints, solving the constraints can be fast as the constraints allow the solver to prune the search space.}. Developers of synthesis systems put significant effort into simplifying and rewriting the constraint formula into a more compact representation ~\citep{rohitRewrite,klee}. Nonetheless, to apply program synthesis to a large dataset, one needs to limit the number of examples expressed as constraints.

The standard method to limit the number of examples is CEGIS (counter example guided inductive synthesis) \citep{sketch06}. CEGIS employs two adversarial sub-routines, a synthesizer and a checker: The synthesizer solves the CSP on a subset of examples rather than on the whole set, producing a candidate program; the checker takes the candidate program and produces an adversarial counter example that invalidates the candidate program. This adversarial example is then added to the subset of examples, prompting the synthesizer to improve. CEGIS successfully terminates when the checker fails to produce an adversarial example. By iteratively adding counter examples to the subset as needed, CEGIS can drastically reduce the size of the constraints constructed by the synthesizer, making it scalable to large datasets. However, CEGIS has to repeatedly invoke the constraint solver in the synthesis sub-routine, solving a sequence of challenging CSP problems. Moreover, due to the phase transition \citep{gent1994sat} property of SAT formulas, there may be instances in the sequence of CSPs with enough constraints to make the problem difficult, yet not enough constraints for the solver to prune the search space\footnote{Imagine a mostly empty Sudoku puzzle, the first few numbers and the last few numbers are easy to fill, whereas the intermediate set of numbers are the most challenging}, making the performance of CEGIS volatile.  

We describe a method that iteratively construct a \emph{representative subset} of examples, which is both sufficient to specify a correct program and small enough to encode efficiently by a constraint solver. The algorithm is greedy: Starting with a (potentially empty) subset of examples, it uses a pre-trained neural network to compute the probability of other examples not in the subset conditioned on the subset, and extends the subset with the most ``surprising'' example (one with the smallest probability). The reason being if an example has a low probability conditioned on the given subset, then it is the most constraining example that can maximally prune the search space once added. The algorithm stops when all the input-output examples have a sufficiently high probability. Experiments show that our method does find representative subsets most of the times, and significantly improves synthesis time and stability on the the tasks of automaton induction and inverse rendering against strong baselines.

\section{An Example Synthesis Problem}

\begin{figure}[!ht]
  \centering
    \textbf{A Drawing Program and its Rendering}\par\medskip
    \includegraphics[scale=0.42]{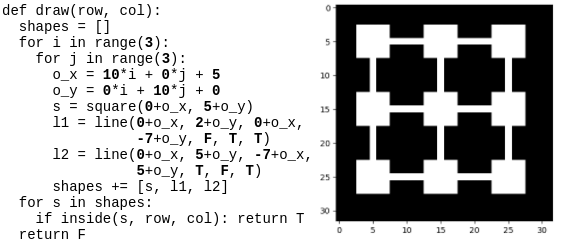}
  \caption{An example draw function (left) and its corresponding rendering (right). The parameters of the draw function are in bold, such as the number of iterations and offsets for the shapes.}
    \label{fig:sketch}
\end{figure}

To best illustrate the synthesis problem and explain our approach, consider a diagram drawing DSL \citep{ellis2017learning} that allows a user to draw squares and lines. The DSL defines a $draw(row, col)$ function, which maps a $(row, col)$ pixel-coordinate to a boolean value indicating whether the specified pixel coordinate is contained within one of the shapes. By calling the $draw$ function across a canvas, one obtains a rendering of the image where a pixel coordinate is colored white if it is contained in one of the shapes, and black otherwise. Figure~\ref{fig:sketch} shows an example of a draw function and its generated rendering on a 32 by 32 pixel grid. The DSL contains a set of parameters that allows the $draw$ function to express different diagrams, which are in bold in Figure~\ref{fig:sketch}(left). The synthesis task is: Given a diagram rendered in pixels, discover the parameter values in the draw function so that it can reproduce the same rendering.

The synthesized drawing program is correct when its rendered image matches the target rendering exactly. As the DSL contains control flow structures such as ``for'' and ``if'', it is a difficult combinatorial problem that requires the use of a constraint solver. Let $Sdraw$ be the synthesized draw function and $Target$ be the target rendering:
\begin{equation*}
\begin{split} 
& \textbf{correct}(Sdraw) \coloneqq \\
& \bigwedge_{(row, col)} Sdraw(row, col) = Target[row][col]
\end{split}
\end{equation*} 
Here, each of the pixels in the target render is encoded as an input-output pair $((row,col), bool)$ that generates a distinct constraint on all of the parameters. For the 32 by 32 pixel image, a conjunction of 1024 distinct constraints are generated, which impose a significant encoding overhead.

\begin{figure}[!ht]
  \centering
    \textbf{Selected Subsets and Synthesized Parameters}\par\medskip
\includegraphics[scale=0.40]{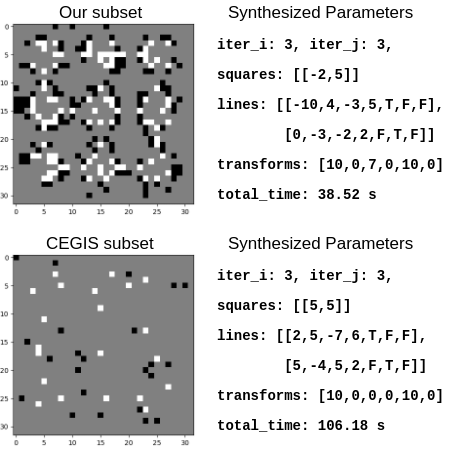}
  \caption{Our algorithm selects more examples than CEGIS, but since the subset is more representative, we achieve better time.}
    \label{fig:selected_pixels}
\end{figure}

In this paper, we propose an algorithm that approximates a representative subset of input-output examples. This subset is small, which alleviates the encoding overhead, yet remains representative of all the examples so that it sufficiently specifies the correctness condition. Figure~\ref{fig:selected_pixels} (top, left) shows the subset chosen by our algorithm. As we can see, from a total of 1024 examples, only 15\% are selected for the representative subset. The representative subset is then given to the constraint solver, recovering the hidden parameter values in Figure~\ref{fig:selected_pixels} (top, right). By comparison, the CEGIS algorithm chooses a much smaller number of examples that are not representative, Figure~\ref{fig:selected_pixels} (bottom, left): Despite the small size of the subset, since it is not representative, CEGIS ultimately achieves a longer solving time.

\begin{figure}[!ht]
  \centering
    \textbf{Iterative Subset Construction}\par\medskip
\includegraphics[scale=0.35]{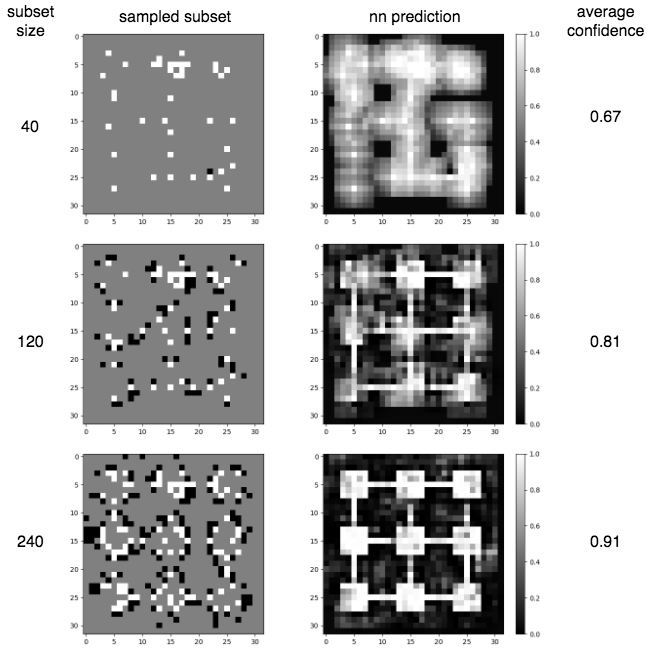}
  \caption{At each step, our algorithm predicts the pixel values of all the pixels conditioned on the sampled pixels, and samples additional pixels with the greatest reconstruction errors.}
    \label{fig:predicted_pixels_graphix}
\end{figure}

Our algorithm constructs the representative subset iteratively. Starting with an empty subset, the algorithm uses a neural network to compute the probability of all the examples conditioned on the chosen examples in the subset. It then adds the least probable example to the subset, the intuition being the example with the lowest probability would best prune the search space as a constraint. The algorithm terminates when all the examples in the dataset are given a sufficiently high probability. An example execution of our selection algorithm is shown in Figure~\ref{fig:predicted_pixels_graphix}. The rest of the paper elaborates our approach.

\section{Discovering Representative Examples}

The crux of our algorithm is an example selection scheme, which takes in a set of examples and outputs a small subset of representative examples.  Let $\subD \subseteq \D$ be a subset of examples. Abusing notation, let us define the \emph{consistency constraint} $\subD(s) \coloneqq \bigwedge_{(x_i,y_i) \in \subD} F(x_i;s)=y_i$, that is to say, the parameter\footnote{We'll refer to $s$ as either a ``parameter'' or a ``program'' from now on, whichever is most appropriate given the context.} $s$ is consistent with all examples in $\subD$. We define the \emph{optimal representative subset} as:
\begin{equation*}
D^* = \argmin_{\subD \subseteq \D} |\subD| ~~s.t.~~ \forall s \in S . ~~ \subD(s) \Rightarrow \D(s) .
\end{equation*} 
$\D^*$ is \emph{representative} of $\D$ in a sense any parameter $s$ satisfying $\D^*$ must also satisfy $\D$. Finding the exact minimum sized $\D^*$ is often intractable, thus we focus on finding a sufficient subset that is as close in size to $\D^*$ as possible.

\subsection{Examples Selection: a Greedy Strategy}

We start with an approximate algorithm with a count oracle $c$, which counts the number of valid solutions with respect to a subset of examples: $c(\subD) \coloneqq |\{s \in S | \subD(s) \}|$. Algorithm \ref{alg:alg_count} constructs the subset $\subD$ greedily, choosing the example that maximally prunes the solution space.
\begin{algorithm}[tb]
   \caption{greedy selection with count oracle $c$}
   \label{alg:alg_count}
\begin{algorithmic}
  \STATE {\bfseries Input:} data $\D$
  \STATE {\bfseries Output:} data subset $\subD$
  \STATE Initialize $\subD = \{\}$.
  \REPEAT
	\STATE $(x,y) \leftarrow \argmin_{(x_j,y_j)} c(\subD \cup \{(x_j,y_j)\})$ \# selection
    \STATE $\subD \leftarrow \subD \cup \{(x,y)\}$
  \UNTIL{$c(\subD) = c(\subD \cup \{(x,y)\})$}
  \STATE {\bfseries return} $\subD$
\end{algorithmic}
\end{algorithm}

\paragraph{Claim 1:} Algorithm \ref{alg:alg_count} produces a subset $\subD$ that is representative, i.e. $\forall s \in S.~ \subD(s) \Rightarrow \D(s)$.
\paragraph{Proof 1:} As $\subD(s)$ is defined as a conjunction of satisfying each example, $c$ can only be monotonically decreasing with each additional example/constraint: $c(\subD) \geq c(\subD \cup \{(x,y)\})$. At termination, the counts remain unchanged $c(\subD) = c(\subD \cup \{(x,y)\}), \forall (x,y) \in \D $, meaning no more solutions can be invalidated. Thus we obtain the sufficiency condition $\forall s \in S.~ \subD(s) \Rightarrow \D(s)$.

\paragraph{Claim 2:} Let $\prune(\D) \coloneqq |\{s \in S | \neg \D(s) \}|$ denotes the number of programs invalidated by $\D$, and $k^{opt} = |\D^*|$ the size of the optimal subset, then the greedy subset returned by Algorithm \ref{alg:alg_count} satisfies $|\greedD| < \frac{\log(\prune(\D))}{\log(k^{opt}) - \log(k^{opt}-1)}$ 

\paragraph{Lemma 2.1:} It will be helpful to first show that the function $\prune(\cdot)$ is both monotonic and sub-modular.

\paragraph{Proof 2.1:} To show monotonicity, note that the constraint generated by the examples are conjunctive, thus adding examples strictly increases the number of invalidated programs. 

To show sub-modularity, we require for $A \subseteq B \subseteq D$:
\begin{equation*}
\begin{split} 
\forall (x, y) \in D . ~ & \prune(A \cup \{(x,y)\}) - \prune(A)\\ 
 \geq ~ & \prune(B \cup \{(x,y)\}) - \prune(B) 
\end{split}
\end{equation*} 
Let $A'(s) \coloneqq A(s) \wedge \neg \{(x,y)\}(s)$, the constraint stating that a program $s$ should satisfy $A$, but fails to satisfy $(x,y)$; Similarly, let $B(s)' \coloneqq B(s) \wedge \neg \{(x,y)\}(s)$. Then, the count $c(A')$ measures how many parameter $s$ becomes invalidated by introducing $(x,y)$ to $A$, i.e. $c(A') = \prune(A \cup \{(x,y)\}) - \prune(A)$, similarly, $c(B') = \prune(B \cup \{(x,y)\}) - \prune(B)$. Note that $A'$ and $B'$ are conjunctive constraints, with $B'$ strictly more constrained than $A'$ due to $A \subseteq B$. Thus $c(A') \geq c(B')$, and we have sub-modularity of $\prune(\cdot)$ as claimed. 

\paragraph{Proof 2:} 

We now derive an upper bound on the size of the representative subset returned by Algorithm \ref{alg:alg_count}. As $\prune(\cdot)$ is monotonic and sub-modular, \citet{nemhauser1978analysis} showed that for any optimal subset of size $k = |\D^*_{k}|$, the greedily constructed subset of size $i = |\greedD_{i}|$ satisfies:
\begin{equation*}
\prune(\greedD_{i}) \geq (1 - (\frac{k-1}{k})^i) \prune(\D^*_{k}) ~.
\end{equation*} 
Let $\rem(\subD) = \prune(D) - \prune(\subD)$ be the remaining number of solutions yet to be pruned by $\subD \subseteq \D$. After subtracting both sides of the inequality from $\prune(D)$:
\begin{equation*}
\rem(\greedD_{i}) \leq \prune(\D) - (1 - (\frac{k-1}{k})^i) \prune(\D^*_{k}) ~.
\end{equation*} 
Set $k = k^{opt}$, the size of the optimal representative subset, we can substitute $\prune(\D^*_{opt})$ with $\prune(\D)$:
\begin{equation*}
\rem(\greedD_{i}) \leq \prune(\D) - (1 - (\frac{k^{opt}-1}{k^{opt}})^i) \prune(\D) ~.
\end{equation*} 
Algorithm \ref{alg:alg_count} terminates when there are no more programs to prune, i.e. when $\rem(\greedD_{i}) < 1$:
\begin{equation*}
\rem(\greedD_{i}) \leq \prune(\D) - (1 - (\frac{k^{opt}-1}{k^{opt}})^i) \prune(\D) < 1 ~.
\end{equation*} 
Rearranging terms we see Algorithm \ref{alg:alg_count} terminates when:
\begin{equation*}
i < \frac{log(\prune(\D))}{\log(k^{opt})- \log(k^{opt}-1)} ~.
\end{equation*} 
Unfortunately, this is a rather loose upper-bound as the difference between $\log(k)$ and $\log(k-1)$ is quite small. However, in some instances it can still be helpful: If $\prune(D) = 1.0e6$ and $k^{opt} = 20$, we have $|\greedD| < 270$, which could be significantly smaller than $|D|$. In the experiment section we explicitly measure the size of the subset returned by our algorithm, showing that in practice one could obtain much smaller subsets than this upper-bound.

The issue with Algorithm \ref{alg:alg_count} is that it requires access to a model counting \citep{Gomes08modelcounting} oracle, which is impractical in practice. We now aim to resolve this issue.

\subsection{Example Selection: by Anticipating New Examples}

We describe an alternative selection criteria that can be approximated efficiently with a neural network. Let's write the selected subset $\subD$ as $\{ (x^{(1)}, y^{(1)}) \dots (x^{(r)}, y^{(r)}) \}$ where $(x^{(j)}, y^{(j)})$ denotes the $j^{th}$ input-output example to be added to $\subD$. We define the \emph{anticipation probability}:
\begin{equation*}
\begin{split} 
Pr((x,y) | \subD) \coloneqq & Pr(F(x;s)=y | \subD(s)) \\
= &Pr(F(x;s)=y | F(x^{(1)};s) = y^{(1)},\\
& \dots, F(x^{(r)};s) = y^{(r)})
\end{split}
\end{equation*} 
Note that $Pr((x,y) | \subD)$ is \textbf{not} a joint distribution on $x$ and $y$, but rather the probability for the event where the parameterized function $F(\cdot~;s)$ maps the input $x$ to $y$, conditioned on the event where $F(\cdot~;s)$ is consistent with all the input-output examples in $\subD$. We claim that one can use $Pr((x,y) | \subD)$ as an alternative to the count oracle $c$.

\paragraph{Claim 3:} Assuming uniform distribution $s \sim unif(S)$: 
\begin{equation*}
\argmin_{(x,y)} c(\subD \cup \{(x,y)\}) = \argmin_{(x,y)} Pr((x,y) | \subD) ~.
\end{equation*} 

\paragraph{Proof 3:} The probability $Pr((x,y) | \subD)$ can be written as a summation over all the possible parameter values for $s$:
\begin{equation*}
\begin{split} 
Pr((x,y) | \subD) \coloneqq & Pr(F(x;s)=y | \subD(s) ) \\
= & \sum_{s \in S} Pr(s|\subD(s))Pr(F(x;s)=y|s) ~~.
\end{split}
\end{equation*} 
Note that under $s \sim unif(S)$, we have:
\begin{equation*}
    Pr(s|\subD(s)) =
    \begin{cases}
      \frac{1}{c(\subD)} & \text{if}\ \subD(s) \\    
      0 & \text{otherwise}\ \\
    \end{cases} ~~.
\end{equation*}
And since $F(\cdot~;s)$ is a function we have:
\begin{equation*}
    Pr(F(x;s)=y|s) =
    \begin{cases}
      1 & \text{if}\ F(x;s)=y \\
      0 & \text{otherwise}\ \\
    \end{cases} ~~.
\end{equation*}
Thus the summation over all $s$ results in:
\begin{equation*}
\begin{split} 
\sum_{s \in S} Pr(s|\subD(s))Pr(F(x;s)=y|s) = \frac{c(\subD \cup \{(x,y)\})}{c(\subD)} ~~.
\end{split}
\end{equation*} 
As $c(\subD)$ is constant under $\argmin_{(x,y)}$ given $\subD$, we have $\argmin_{(x,y)} c(\subD \cup \{(x,y)\}) = \argmin_{(x,y)} Pr((x,y) | \subD)$ as claimed.

It is easy to see that one needs to update the termination condition to $\min_{(x,y)} Pr((x,y) | \subD) = 1$, when all the input-output examples are completely anticipated given $\subD$.

\section{Neural Network Model}
We now describe the high level neural network architecture that models the anticipation probability, $Pr((x,y)|\subD)$. 

\subsection{Factorization: Enabling Scaling with $|\subD|$}
A challenge to our neural network encoding is the ability of our model to scale with the size of $\subD$: as we might collect a large subset of examples, and the probability of each new example $(x,y)$ depends on the entire subset $\subD$.  

To address this, we make an independence assumption: Let $\neib_{k,x}$ be the neighborhood function that computes the top-k neighbors of $x$ from $\subD$ and condition $(x,y)$ only on these top-k neighbors:
\begin{equation*}
\begin{split}
Pr((x,y)|\subD) = & Pr((x,y)|\neib_{k,x}(\subD)) \\
                = & Pr((x,y)|(x_{nb}^1, y_{nb}^1), \dots, (x_{nb}^k, y_{nb}^k)) ~.\\
\end{split}  
\end{equation*} 
We assume the programmer would be able to come up with an appropriate neighborhood function for each synthesis task. In our experiments, the neighborhood is measured by a distance metric on the input space $X$. For example, we have used a convolutional neural network (which implicitly uses pixel to pixel distance) and longest matching suffix (sub-string distance). Although we remark that in general, a neighborhood need not depend on a distance metric but can be as arbitrary as needed. 

\subsection{Anticipation Network: Direct Computation}
\begin{figure}[!ht]
  \centering
  \textbf{Neural Network Architecture}\par\medskip
    \includegraphics[scale=0.32]{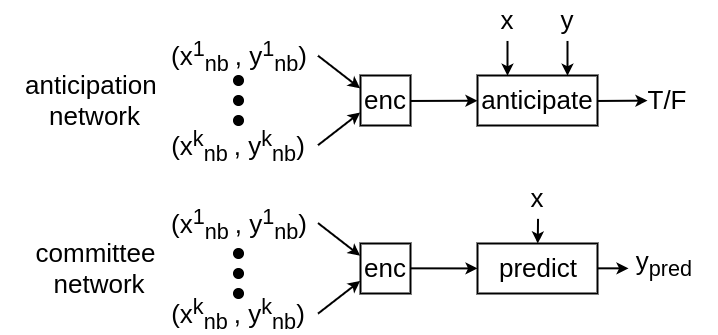}
      \caption{The anticipation network (top) that computes the anticipation probability directly, and the committee network (bot) computes the output on a new given input $x$}
    \label{fig:nn_arch}
\end{figure}

Figure \ref{fig:nn_arch} (top) shows a neural network architecture that models the factorized anticipation probability $Pr((x,y)|(x_{nb}^1, y_{nb}^1), \dots, (x_{nb}^k, y_{nb}^k))$ directly. 

We train the network on the task of correctly anticipating whether an input-output pair $(x,y)$ should occur based on the top-k neighbors of $(x,y)$ from a subset $\subD$. To do this, we sample a program $s \sim unif(S)$ and a subset of inputs $\{x_1 \dots x_n | x_i \sim unif(X) \}$. We evaluate the program on the set of inputs to produce a dataset of example pairs $\D = \{(x_1, F(x_1;s)) \dots (x_n, F(x_n;s))\}$. We can then sample a subset $\subD \subseteq \D$ and a new example $(x,y) \in \D \setminus \subD$. From this new example, we compute its top-k neighbors $\{(x_{nb}^1, y_{nb}^1), \dots, (x_{nb}^k, y_{nb}^k)\} = \neib_{k,x}(\subD)$. We also construct a negative sample $(x,y_{neg})$ by sampling a random $y_{neg} \in Y \setminus \{y\}$. The network is then trained to produce $True$ on the input $\{(x_{nb}^1, y_{nb}^1), \dots, (x_{nb}^k, y_{nb}^k), (x,y)\}$ and to produce $False$ on the input $\{(x_{nb}^1, y_{nb}^1), \dots, (x_{nb}^k, y_{nb}^k), (x,y_{neg})\}$

\subsection{Committee Network: Peer Consultation}
We now present an equivalent neural network architecture that affords a more intuitive understanding. Figure \ref{fig:nn_arch} (bot) shows the committee network, which computes an output distribution $y_{pred}$ rather than the anticipation probability. The two architectures are equivalent because:
\begin{equation*}
\begin{split}
  &Pr((x,y)|(x_{nb}^1, y_{nb}^1), \dots, (x_{nb}^k, y_{nb}^k)) \\
= &Pr(F(x;s)=y|F(x_{nb}^1;s)=y_{nb}^1, \dots, F(x_{nb}^k;s)=y_{nb}^k)
\end{split}  
\end{equation*} 
The committee network is trained on the task of producing the correct value $y$, which has the implicit effect of negative sampling. This network has a very intuitive interpretation: To best predict the a function's output on a new example $x$, we consult the subset $\subD$ for the top-k most relevant input-output pairs to make a prediction on the value of $y$.

In practice, each synthesis domain would require a different neural-network architecture, as the input/output types of the functions being synthesized and the neighborhood function are domain specific. However, the overall neural-network task remains the same: predicting the function's output on a new example $x$ based on the nearest-k neighbors of $x$ already present in the subset $\subD$. We'll describe the domain specific architecture in detail in the Experiment section.
\section{Synthesis with Representative Examples}
The neural network cannot perfectly model the anticipation probability, thus, our example selection algorithm can only approximate a representative subset, causing the synthesized program to be inconsistent with the entire dataset of examples. We remedy this problem by combining example selection and CEGIS, getting the best of both worlds. 

\subsection{CEGIS: Guarantees with Caveats}
CEGIS \citep{sketch06} is a synthesis algorithm which guarantees total correctness on a set of examples $\D$. It is outlined in Algorithm \ref{alg:alg_cegis}. CEGIS is composed of two adversarial sub-routines, a synthesizer and a checker: The synthesizer produces a candidate program $s$ over the subset $\subD$, which is initially empty; The checker takes in this candidate program $s$ and produces an adversarial counter example $ce \in \D \setminus \subD$ which invalidates $s$. $ce$ is added to the subset of examples, prompting the synthesizer to improve. CEGIS successfully terminates when the checker fails to produce an adversarial example.
\begin{algorithm}[tb]
   \caption{CEGIS}
   \label{alg:alg_cegis}
\begin{algorithmic}
  \STATE {\bfseries Input:} data $\D$, initial subset $\subD = \{\}$
  \STATE {\bfseries Output:} satisfying program $s$
  \REPEAT
	\STATE $s \leftarrow synthesize(\subD)$
    \STATE $ce \leftarrow check(s,\D)$
    \STATE $\subD \leftarrow \subD \cup \{ce\}$
  \UNTIL{$check(s, \D) =$ None}
  \STATE {\bfseries return} $s$
\end{algorithmic}
\end{algorithm}

At first glance CEGIS is very similar to our approach, but a deeper look reveals several important differences. First, the subset $\subD$ of examples constructed by CEGIS upon termination is \emph{not} representative: It is possible for CEGIS to synthesize the correct program without constructing a representative subset by luck, a fact we demonstrate empirically in the experiments. The danger of synthesis over a non-representative subset is that there might be instances where there are enough constraints to make the synthesis problem challenging, yet not enough constraints for the solver to prune the search space. The result is the hanging of the solver for an extended periods of time, without any guarantee whether the synthesis would ever terminate. Secondly, to build the subset of counter examples, CEGIS must solve $|\subD|$ instances of constraint problems, each one with the potential to timeout due to being under-constrained.

\subsection{Our Algorithm: Best of Both Worlds}
Our algorithm combines representative example discovery and CEGIS by instantiating the subset of counter examples in CEGIS with a representative subset, see Algorithm \ref{alg:alg_ours}. This algorithm guarantees complete correctness over the input dataset $\D$ while alleviating the challenges of CEGIS by presenting CEGIS with a well-constrained subset upfront. 

\begin{algorithm}[tb]
   \caption{Synthesis with Representative Examples}
   \label{alg:alg_ours}
\begin{algorithmic}
  \STATE {\bfseries Require:} trained committee model $nn(\neib_{x,k}(\subD), x)$\\
  approximating $Pr(y | \neib_{x,k}(\subD), x)$
  \STATE {\bfseries Input:} data $\D$
  \STATE {\bfseries Output:} satisfying program $s$
  \STATE Initialize $\subD = \{\}$.
  \REPEAT
  	\STATE \# Find the least likely input-output example
	\STATE $(x,y) \leftarrow \argmin_{(x_j,y_j)} nn(\neib_{x_j,k}(\subD), x_j)(y_j)$
    \STATE $\subD \leftarrow \subD \cup \{(x,y)\}$
  \UNTIL{confident$(nn, \subD, \D)$}
  \STATE {\bfseries return} CEGIS$(\D, \subD)$
\end{algorithmic}
\end{algorithm}

\section{Experiments}

Our approach is evaluated against two criteria: First, the representativeness of our selected subset is explicitly measured; Then, the time/stability improvement of using such a subset is measured against several strong baselines.

\subsection{Explicitly Measuring Representativeness}
This experiment explicitly measures the representativeness of the subset selected by our algorithm on the task of ordering synthesis: Given a dataset of pair-wise ordering relations, $\D = \{a<b, a<c, b<d, c<d, d>a, c>a\}$, the task is to synthesize any total-ordering that is consistent with $\D$, for instance, $(a,b,c,d)$ or $(a,c,b,d)$. This task is useful because the optimal representative subset can be constructed as a Hasse diagram \citep{ullmantransitive} by pruning transitive relations: $\D^* = \{a<b, a<c, b<d, c<d\}$. Thus, we can measure both the representativeness and optimality of our selection algorithm. In this experiment, we set $n=10$ and give our selection algorithm a dataset of size $30\%$ to $100\%$ of all possible pair-wise orderings. Since there are only $100$ possible pair-wise relations for $n=10$, we use a fully-connected neural network without factorization. Refer to the supplementary for  specifics of this network and an algorithm to verify representativeness. 

\begin{figure}[!ht]
  \centering
  \textbf{Subset sizes and representativeness}\par\medskip
  \includegraphics[scale=0.37]{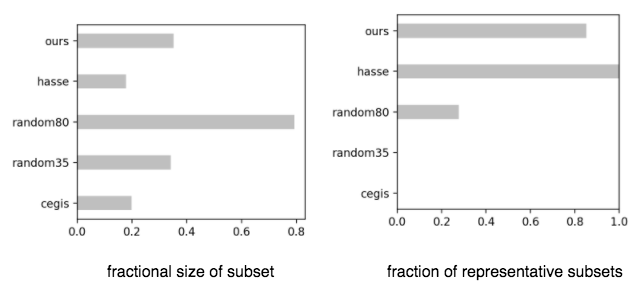}
  \caption{Our approach discovers representative subsets 85\% of the times while sampling $2\times$ the optimal subset size. Measured on 500 datasets drawn from randomly sampled total orderings}
    \label{fig:experiments_representative}
\end{figure}

\begin{figure}[!ht]
  \centering
  \textbf{Chosen Subsets on a Particular Dataset}\par\medskip
    \includegraphics[scale=0.22]{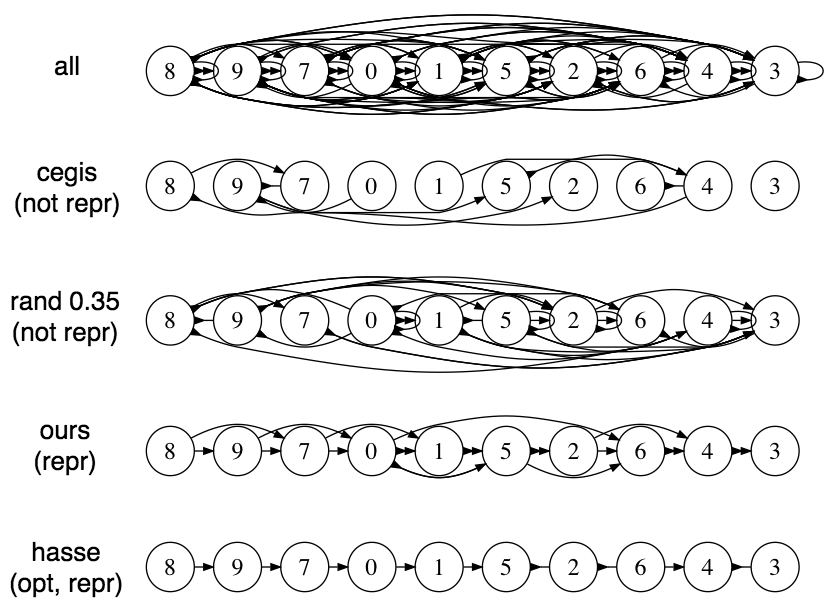}
  \caption{Chosen subsets on a particular dataset \textbf{all}. A subset is representative if it contains all adjacent pair-wise ordering}
    \label{fig:experiments_representative_vis}
\end{figure}

We test the representativeness of our approach against the following baselines: \textbf{cegis}, \textbf{random x}(randomly select x percent of dataset)\footnote{we use 35\% because it matches our average subset size}, and \textbf{hasse} (the optimal construction). The measurement of average subset size and fraction of representative subsets is shown in Figure~\ref{fig:experiments_representative}. As we can see, \textbf{our} approach selects about twice as many examples as the optimal subset, and 85\% of the times our subsets are representative. By contrast, \textbf{cegis} and \textbf{rand35} fails to discover any representative subset, while \textbf{rand80} discovers representative subset only 30\% of the times despite sampling 80\% of the total data. Figure~\ref{fig:experiments_representative_vis} visualizes the chosen pair-wise orderings on a particular dataset \textbf{all}. This dataset specifies a unique total-ordering, which \textbf{hasse} was able to concisely represent with the minimal representative subset (bottom). \textbf{our} approach also discovers a representative subset, albeit with a few extra redundant relations. By contrast, \textbf{cegis} and \textbf{rand35} fail to discover a representative subset, as their subsets lack the relationship between elements $7$ and $0$, which are adjacent in the total ordering.

\subsection{Measuring Improved Synthesis Times}
We now measure the performance in terms of on time and stability by using our approach on two distinct tasks.

\paragraph{DFA Synthesis} The task is to synthesize a deterministic finite-state automaton (DFA) from a set of accepted and rejected strings. We use a DSL which contains DFA of $6$ states over a binary alphabet of $0$ and $1$ with a single accept state. The search space of total possible DFAs is of size $6^{12} = 2.18 \times 10^9$. 1000 strings of variable length between 5 and 10 were provided as the dataset for each synthesis task, the experiment consists of $400$ tasks. On this domain, given a new example string $str$, the neighborhood function selects the top $10$ closest prefix and suffix matching examples from the subset $\subD$. The neural network architecture is a simple feed-forward neural network that predicts the accept/reject label of $str$ directly (see supplementary for parameters details). 
\begin{figure}[!ht]
  \centering
  \textbf{Automaton Synthesis}\par\medskip
  \includegraphics[scale=0.37]{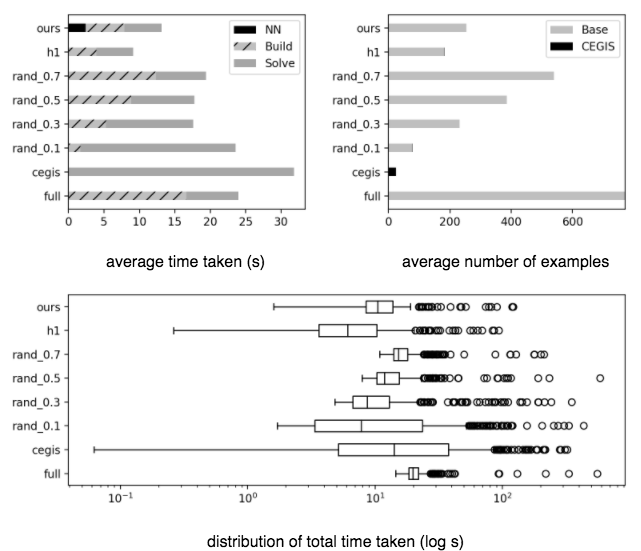}
  \caption{Time performance on DFA synthesis. \textbf{our} approach nearly matches the crafted heuristic \textbf{h1}, which constructs a suffix-tree over the entire dataset $\D$, and out performs all other baselines.}
    \label{fig:experimentsdfa}
\end{figure}

We measure performance against the following: \textbf{full} (all examples are added), \textbf{cegis},\textbf{rand\_x} (initialize CEGIS with a random $x$ fraction of data), \textbf{h1} (a heuristic that construct a suffix-tree over the entire dataset, see supplementary). Figure~\ref{fig:experimentsdfa} (top,left) shows the comparison of performances in average time. As we can see, the heuristic subset collection \textbf{h1} performs best on average, but \textbf{our} approach comes in close (If we disregard the example selection time from NN, the two performs similarly). As we can see, \textbf{ours}, \textbf{h1}, \textbf{full} have similar solve time, which we can infer that \textbf{our} approach and \textbf{h1} have found a well-constraining representative subset. This is in stark contrast to \textbf{cegis} which explodes in solve time with hardly any examples chosen. In terms of stability (Figure~\ref{fig:experimentsdfa} (bot)), \textbf{our} approach also closely matches that of the heuristic, whereas all other algorithms (except \textbf{full}) suffers big variance in total time, likely a result of performing synthesis on under-representative subsets. Figure~\ref{fig:experimentsdfa} (top,right) shows the average number of examples in the collected subset, we see that \textbf{our} approach outperforms randomly selected subsets of any size.

\paragraph{Programmatic Drawing Synthesis} We evaluate our approach on 250 randomly sampled 32$\times$32 pixel renderings created from the drawing DSL in Section 2, the drawing function has a parameter space of size $1.31\times10^{23}$. The neighborhood function is simply a $7 \times 7$ sliding window centered on each pixel, and is implemented as a convolutional neural network (see supplementary for parameter details).

\begin{figure}[!ht]
  \centering
  \textbf{Programmatic Drawing Synthesis}\par\medskip
\includegraphics[scale=0.37]{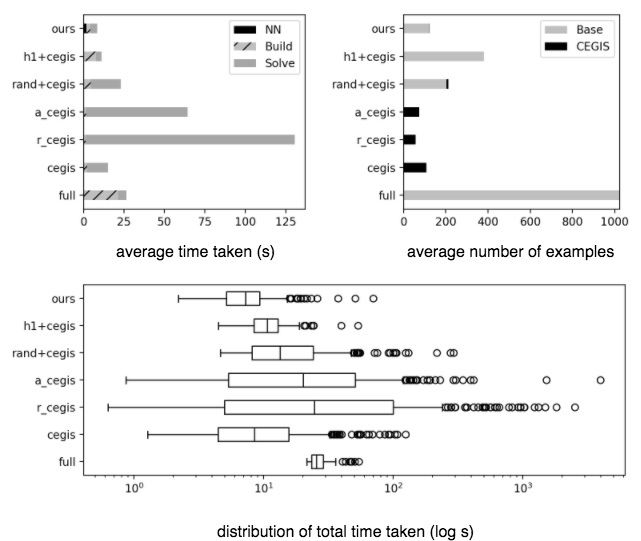}

  \caption{Time performance on programmatic drawing synthesis. \textbf{our} approach is best in average time, and achieves similar stability as \textbf{full} and \textbf{h1+cegis} with much fewer samples.}
    \label{fig:experimentsgraphix}
\end{figure}

We measure performance against the following: \textbf{full} (all examples are added), \textbf{cegis, rcegis, acegis} (different CEGIS flavours on how the counter examples are selected: canonical top-left most pixel, random, and a fixed but arbitrary order), \textbf{rand+cegis} (instantiate CEGIS with a random 20\% subset), and \textbf{h1+cegis} (a heuristic that adds a pixel if any pixel within a $5 \times 5$ window has a different value). The results are shown in Figure~\ref{fig:experimentsgraphix}. As we can see, \textbf{our} approach performs best on average, beating all competitors on average time. One unexpected outcome is that \textbf{cegis} performs very well on this domain: We postulate that the top-left-most counter-examples chosen by \textbf{cegis} happen to be representative as they tend to lay on the boundaries of the shapes, which is well suited for the drawing DSL domain. However, such coincidence is not to be expected in general: By making the counter example be given at random, or given at a fixed but arbitrary ordering, \textbf{rcegis} and \textbf{acegis} were unable to pick a representative set of examples and suffer in overall time. In terms of variance (Figure~\ref{fig:experimentsgraphix} (bot)), our approach was able to match the variance of \textbf{full} (clear representative) and \textbf{h1+cegis} (also representative as a $5 \times 5$ sliding window can distinguish squares and lines perfectly). However, \textbf{our} approach was able to discover representative subsets with a much smaller number of examples (Figure~\ref{fig:experimentsgraphix} (top, right)).

Overall, our approach improves synthesis time and stability by providing CEGIS with a representative subset upfront. \footnote{The supplementary material and the code can be found at \url{https://github.com/evanthebouncy/icml2018_selecting_representative_examples} }
\section{Related Work}
In recent years there have been an increased interest in \emph{program induction}. \citet{graves2014neural}, \citet{reed2015neural}, \citet{neelakantan2015neural} assume a differentiable programming model and learn the operations of the program end-to-end using gradient descent. In contrast, in our work we assume a non-differentiable programming model, allowing us to use expressive program constructs without having to define their differentiable counter parts. Works such as \citep{reed2015neural} and \citep{cai2017making} assume strong supervision in the form of complete execution traces, specifying a sequence of exact instructions to execute, while in our work we only assume labeled input-output pairs to the program, without any trace information.

\citet{parisotto2016neuro} and \citet{balog2016deepcoder} learn relationships between the input-output examples and the structures of the program that generated these examples. When given a set of input-outputs, these approaches use the learned relationships to prune the search space by restricting the syntactic forms of the candidate programs. In contrast, our committee network learns a relationship between the input-output examples, a relationship entirely in the semantic domain. In this sense, these approaches are complimentary.

The predictive task of our neural network is similar to that of \citep{pu2017acquire}, which learns the inter-relationships between observations for active learning. In contrast, in our domain the labels to the observations are known in advance. The committee neural-network structure is most similar to the meta program-induction network in \citep{devlin2017neural}. One key difference being we assume a neighborhood function which both limits and orders neighboring input-output examples to be encoded. As our subset $\subD$ can grow arbitrarily large, having a hard cap on the number of neighbors is important for efficiency. 
\section*{Acknowledgements}
We like to thank the reviewer for their helpful insights; Xin Zhang, Osbert Bastani for constructive criticisms; Steve Mussmann for in depth reference on sub-modularity; and Twitch Chat for moral supports.

This work was funded by the MUSE program (Darpa grant
FA8750-14-2-0242).

\bibliography{icml_2018}
\bibliographystyle{icml2018}

\end{document}